\documentclass[10pt,twocolumn,letterpaper]{article}

\usepackage{cvpr}

\usepackage{graphicx}
\usepackage{amsmath}
\usepackage{amssymb}
\usepackage{booktabs}

\usepackage[pagebackref,breaklinks,colorlinks]{hyperref}

\usepackage[capitalize]{cleveref}
\crefname{section}{Sec.}{Secs.}
\Crefname{section}{Section}{Sections}
\Crefname{table}{Table}{Tables}
\crefname{table}{Tab.}{Tabs.}

\def\cvprPaperID{1238} % *** Enter the CVPR Paper ID here
\def\confName{CVPR}
\def\confYear{2023}

\newcommand{\ourproposedmethod}{MoDAR}

\usepackage{booktabs}       % professional-quality tables
\usepackage{multirow}
\usepackage{pifont}% http://ctan.org/pkg/pifont
\usepackage[table]{xcolor}
\usepackage{array}
\newcolumntype{H}{>{\setbox0=\hbox\bgroup}c<{\egroup}@{}}
\newcommand{\cmark}{\ding{51}}%
\newcommand{\xmark}{\ding{55}}%
\definecolor{Highlight}{HTML}{39b54a}
\newcommand{\improves}[1]{{ \color{Highlight} {(#1)}}}

\makeatletter
\newcommand{\ssymbol}[1]{^{\@fnsymbol{#1}}}
\def\blfootnote{\xdef\@thefnmark{}\@footnotetext}
\newcommand{\printfnsymbol}[1]{%
  \textsuperscript{\@fnsymbol{#1}}%
}
\makeatother

\begin{document}

\title{MoDAR: Using Motion Forecasting for 3D Object Detection \\in Point Cloud Sequences}

\author{Yingwei Li\thanks{equal contributions}~~~~Charles R. Qi\printfnsymbol{1}~~~~Yin Zhou~~~~Chenxi Liu~~~~ Dragomir Anguelov \\ Waymo LLC}
\maketitle

%%%%%%%%% ABSTRACT
\begin{abstract}
% \todo{Rewrite the motivation part}
Occluded and long-range objects are ubiquitous and challenging for 3D object detection.
Point cloud sequence data provide unique opportunities to improve such cases, as an occluded or distant object can be observed from different viewpoints or gets better visibility over time.
However, the efficiency and effectiveness in encoding long-term sequence data can still be improved.
In this work, we propose MoDAR, using motion forecasting outputs as a type of virtual modality, to augment LiDAR point clouds. The MoDAR modality propagates object information from temporal contexts to a target frame, represented as a set of virtual points, one for each object from a waypoint on a forecasted trajectory. 
A fused point cloud of both raw sensor points and the virtual points can then be fed to any off-the-shelf point-cloud based 3D object detector. 
Evaluated on the Waymo Open Dataset, our method significantly improves prior art detectors by using motion forecasting from extra-long sequences (e.g. 18 seconds), achieving new state of the arts, while not adding much computation overhead.
\end{abstract}

%%%%%%%%% BODY TEXT
\section{Introduction}
% \todo{Rewrite the motivation part}

3D object detection is a fundamental task for many applications such as autonomous driving. While there has been tremendous progress in architecture design and LiDAR-camera sensor fusion, occluded and long-range object detection remains a challenge.
Point cloud sequence data provide unique opportunities to improve such cases.
In a dynamic scene, as the ego-agent and other objects move, the sequence data can capture different viewpoints of objects or improve their visibility over time.
The key challenge though, is how to efficiently and effectively leverage sequence data for 3D object detection.

\begin{figure}[t]
\vskip -1ex
\centering
\includegraphics[width=0.8\linewidth]{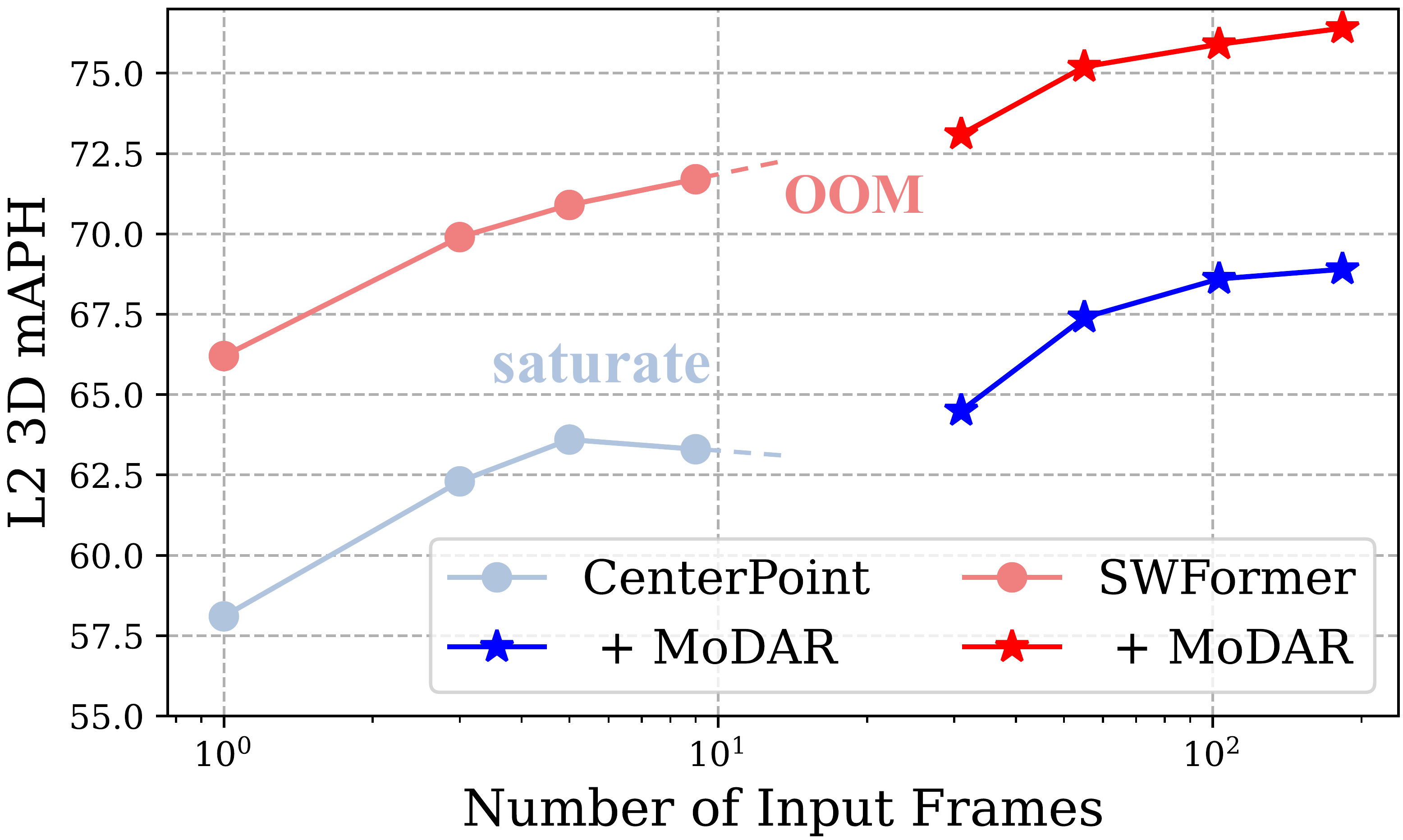}
\vskip -1ex
\captionof{figure}{\textbf{3D detection model performance \textit{vs.} number of input frames.} Naively adding more frames to existing methods, such as CenterPoint~\cite{yin2021center} and SWFormer~\cite{sun2022swformer}, quickly plateaus the gains while our method, MoDAR, scales up to many more frames and gets much larger gains. L2 3D mAPH is computed by averaging vehicle and pedestrian L2 3D APH.}
\label{fig:FrameVSmAPH}
\vskip -4ex
\end{figure}

Existing multi-frame 3D object detection methods often fuse sequence data at two different levels. At scene level, the most straightforward approach is to transform point clouds of different frames to a target frame using known ego motion poses~\cite{caesar2020nuscenes,yang20213d,yin2021center,sun2022swformer}. Each point can be decorated with an extra time channel to indicate which frame it is from. 
However, according to previous studies~\cite{qi2021offboard, chen2022mppnet} and our experiments shown in \cref{fig:FrameVSmAPH}, it is difficult to further improve the detection model by including more input frames due to its large computation overhead as well as ineffective temporal data fusion at scene level (especially for moving objects).
On the other side, 3D Auto Labeling~\cite{qi2021offboard} and MPPNet~\cite{chen2022mppnet} propose to aggregate longer temporal contexts at object level, which is more tractable as there are much less points from objects than those from the entire scenes. However, they also fail to scale up temporal context aggregation to long sequences due to efficiency issues or alignment challenges.

\begin{figure}[t]
\centering
\includegraphics[width=0.85\linewidth]{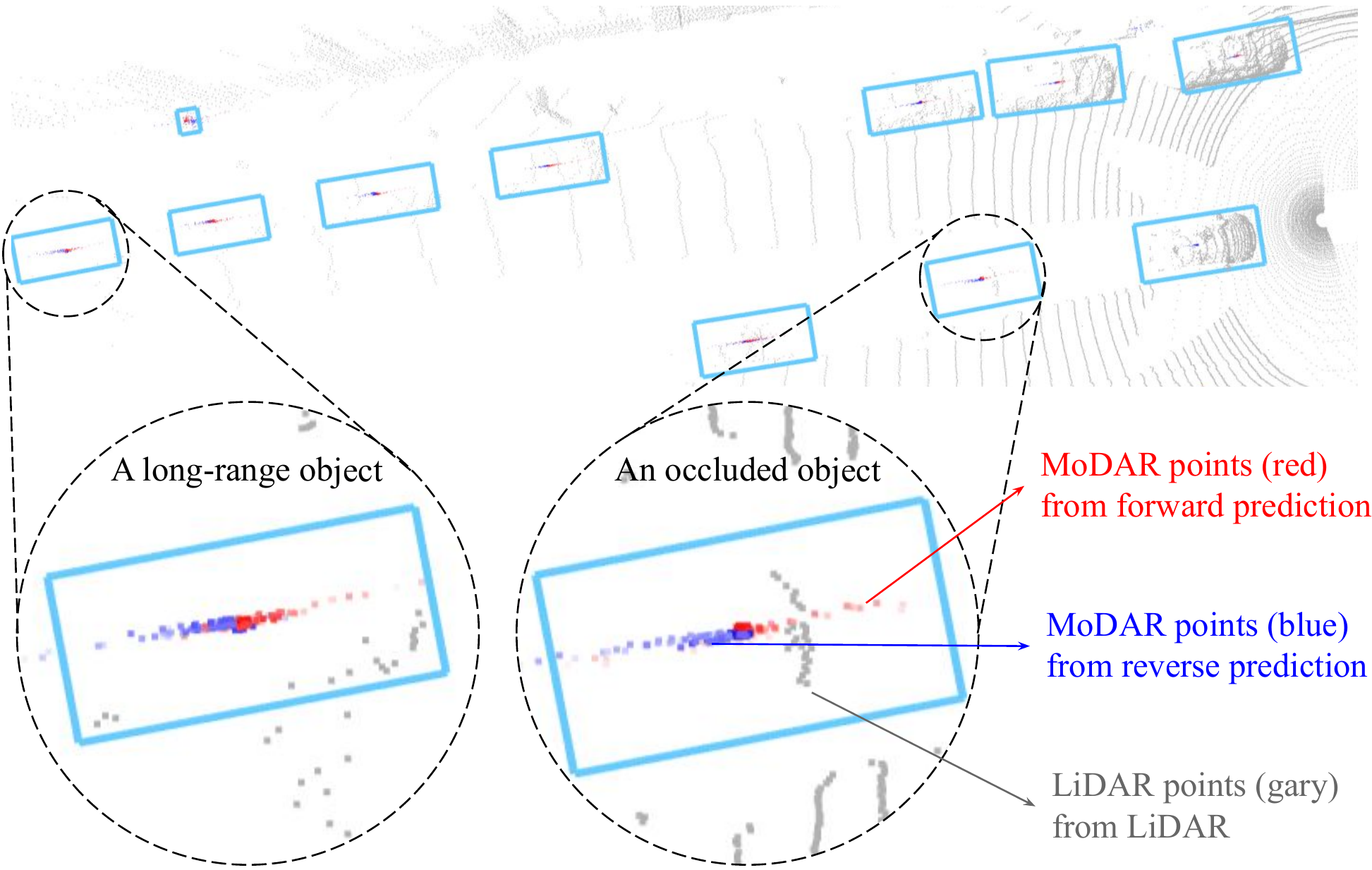}
\vskip -1ex
\caption{\textbf{3D object detection from MoDAR and LiDAR points.} MoDAR points (\textcolor{red}{red} and \textcolor{blue}{blue}) are predicted object centers with extra features such as sizes, semantic classes and confidence scores. Compared to LiDAR-only detectors, a multi-modal detector taking both LiDAR (\textcolor{gray}{gray}) and MoDAR points can accurately recognize occluded and long-range objects that have few observed points.} 
\label{fig:teaser}
\vskip -4ex
\end{figure}

In our paper, we propose to use motion forecasting to propagate object information from the past (and the future) to a target frame. The output of the forecasting model can be considered another (virtual) sensor modality to the detector model. Inspired by the naming of the LiDAR sensor, we name this new modality \emph{MoDAR}, Motion forecasting based Detection And Ranging (see \cref{fig:teaser} for an example).

Traditionally 3D object detection is a pre-processing step for a motion forecasting model, where the detector boxes are either used as input (for past frames) or learning targets (for future frames). In contrast, we use motion forecasting outputs as input to LiDAR-MoDAR multi-modal 3D object detectors.
There are two major benefits of using a MoDAR sensor for 3D object detection from sequence data. First, motion forecasting can easily transform object information across very distant frames (8 seconds or longer). Such propagation is especially robust to occlusions as the forecasting models do not assume successful tracking for trajectory forecasting. Second, considering forecasting output as another sensor data source for 3D detection, it is a lightweight sensor modality, making long-term sequence data processing possible without much computation overhead.

Specifically, in MoDAR, we represent motion forecasting output at the target frame as a set of virtual points (named as MoDAR points), one for each object from a waypoint on a forecasted trajectory. The predicted object location is the 3D coordinate of the virtual point, while additional information (such as object type, size, predicted heading, and confidence score) is encoded into the virtual point features. Each virtual point is appended with a time channel to indicate the context frames it uses for the motion forecasting. For a target frame, we can use forecasted outputs from multiple context frames easily through a union of corresponding virtual points. In an offboard/offline detection setup, we can use both forward prediction and reverse prediction (use future frames as input to the forecasting model) to combine information from the past and the future. For detection, we fuse the raw sensor points (from LiDARs) and the virtual points (from forecasting), and feed them to any off-the-shelf point cloud based 3D detector. 

In experiments, we use a MultiPath++~\cite{varadarajan2022multipath++} motion forecasting model trained on the Waymo Open Motion Dataset~\cite{ettinger2021large} to generate MoDAR points from past 9 seconds for online detection; and from past and future 18 seconds for offline detection. With minimum changes, we adapt CenterPoints~\cite{yin2021center} and SWFormer~\cite{sun2022swformer} detectors for LiDAR-MoDAR 3D object detection.~\footnote{Although we experiment with point-cloud based detectors, MoDAR can be fused with perspective view or camera or radar based detectors as using MoDAR can be considered as a sensor fusion process.} Evaluated on the Waymo Open Dataset~\cite{sun2020scalability}, we show that adding MoDAR significantly improves detection quality, improving CenterPoints and SWFormer by $11.1$ and $8.5$ mAPH respectively; and it especially helps detection of long-range and occluded objects. Using MoDAR with a 3-frame SWFormer detector, we have achieved state-of-the-art mAPH on the Waymo Open Dataset. We further provide extensive ablations and analysis experiments to validate our designs and show impacts of different MoDAR choices.

\section{Related Work}
\noindent\textbf{3D object detection on point clouds.}
Most work focuses on using single-frame input. They can be categorized to methods using different representations such as voxels or pillars~\cite{REF:VotingforVoting_RSS2015,REF:Vote3Deep_ICRA2017,REF:3DFCN_RSJ2017,song2016deep,REF:ku2018joint,REF:yang2018pixor,simony2018complex,zhou2018voxelnet,REF:second_2018,lang2019pointpillars,REF:HVNet2020,REF:PillarNet_ECCV2020}, point clouds~\cite{shi2018pointrcnn,REF:yang2018ipod,REF:StarNet_2019,qi2019deep,yang20203dssd,REF:Point-GNN_CVPR2020}, range images~\cite{REF:VeloFCN2016,REF:lasernet_CVPR2019,REF:bewley2020range, sun2021rsn},~\etc. Liu et al.~\cite{liu2022lidarnas} did a review to put those methods in a unified framework.
Among those methods, CenterPoint~\cite{yin2021center} using anchor-free detection heads~\cite{zhou2019objects} becomes one of the popular single-stage 3D detectors.
% Due to the sparsity of the point cloud, a somewhat orthogonal direction is to develop more efficient approach by ignoring the area without lidar points
On the other hand, more recent methods explore to use transformers for 3D detection~\cite{fan2022embracing, sun2022swformer}. For example, SWFormer~\cite{sun2022swformer} used sparse window based transformers to achieve new state-of-the-art performance. In this work we use these two representative detector architectures for our experiments.

\begin{figure*}[ht!]
\centering
\includegraphics[width=\linewidth]{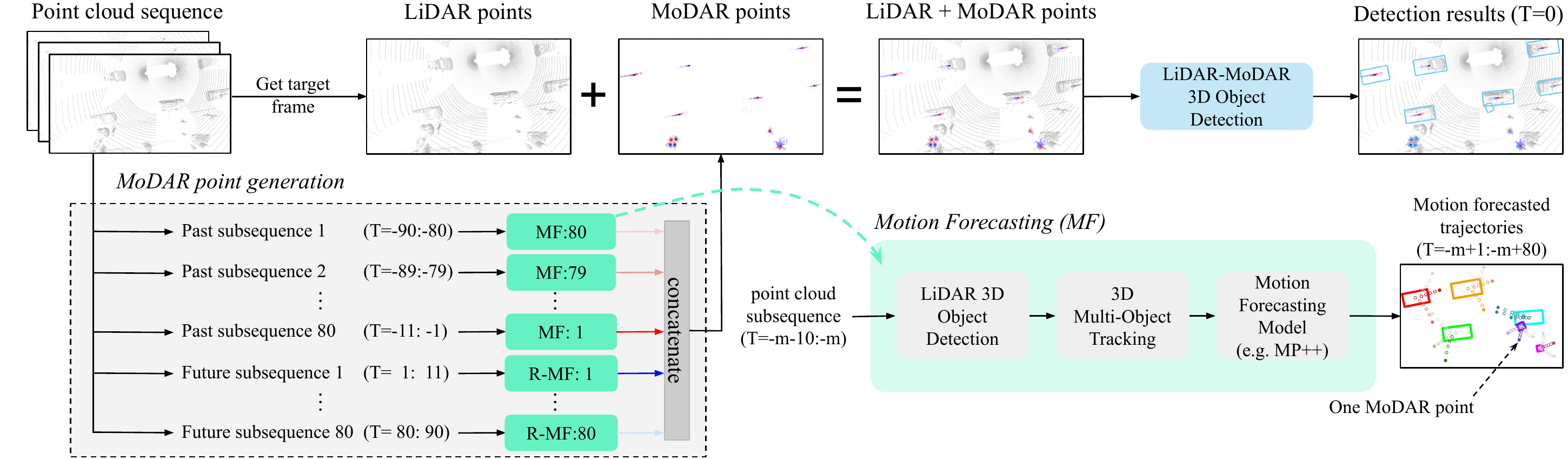}
\caption{\textbf{Using MoDAR for 3D object detection}. Given a point cloud sequence around a target frame, our goal is to estimate 3D object bounding boxes at the target frame T=0. To generate MoDAR points, we run motion forecasting (MF) on various subsequences of the input. MF:m means given a subsequence of frames T=-m-10:-m (11 frames), we run motion forecasting to predict object locations m frames ahead of the subsequence, i.e. predict object location and states at frame T=0. R-MF means reverse motion forecasting, which is only used in the offline use case. The bottom right figure shows how motion forecasting typically works -- it involves running a pre-trained point cloud based detector, a multi-object tracker and a pre-trained motion forecasting model (output colors to indicate different instances). The motion forecasting outputs at the target frame (T=0) from various subsequences are concatenated to form the final set of MoDAR points (with the color indicating from which frame they are predicted). We then union the LiDAR and MoDAR points and train/use a LiDAR-MoDAR multi-modal 3D detector (can be any off-the-shelf 3D detector) to get final detection results at the target frame.}
\label{fig:pipeline}
\vskip -1em
\end{figure*}

\noindent\textbf{Multi-frame 3D object detection.} 
Early multi-frame 3D detectors aggregate features from different frames using convolution layers~\cite{luo2018fast}. More recent methods use a simple point concatenation strategy, which transform short-term point cloud sequences into the same coordinate (using ego-car poses) and then feed the merged points to deep networks~\cite{li2022deepfusion,sun2022swformer,yin2021center,sun2021rsn,ge2020afdet}. They usually use point cloud sequences that are up to 5 frames due to memory/computation costs. Another drawback of point concatenation is that it cannot align moving objects. Later methods explore how to use more frames and model alignment at the intermediate feature level. For example, 3D-MAN~\cite{yang20213d} uses an attention module to align different frames while MPPNet~\cite{chen2022mppnet} designs both intra-group feature mixing and inter-group feature attention. However, these methods are difficult to scale up to more frames due to their large computation overhead. On the other hand, recent methods take bounding boxes from all frames and points from a small set of context frames (for moving objects) as input, but not explicitly handling the alignment issue~\cite{qi2021offboard,yang2021auto4d}.

\noindent\textbf{Multi-modality fusion for 3D object detection.} A robotics system (such as an autonomous driving car) often has multiple sensors, such as LiDARs, cameras, and radars, which provide complementary information. LiDAR-camera fusion is arguably the most common and well-studied modality fusion configuration~\cite{li2022deepfusion,qi2018frustum,vora2020pointpainting,wang2021pointaugmenting,piergiovanni20214d}. 
There is also work on camera-radar fusion~\cite{hwang2022cramnet}. 
Fusion methods generally combine information at input level (early fusion), feature level, or decision level (late fusion)~\cite{pohl1998review}. Input level fusion is usually computationally cheap but needs good cross modality alignment~\cite{vora2020pointpainting}. Specifically, people usually convert camera images to virtual points and fuse virtual points and lidar points as the input~\cite{yin2021multimodal,wu2022sparse}. On the other extreme, decision level fusion can tolerate modality misalignment, but has to a large compute cost or sub-optimal performance~\cite{chatzis1999multimodal}. 
Our work considers motion forecasting as an additional modality for 3D detection, which provides complementary (temporal) information to LiDAR. By carefully designing the format of MoDAR, it aligns well to LiDAR and can be exploited using the efficient input level fusion strategies. Our proposed MoDAR point can be considered as a variant of virtual point, but different from previous works, our MoDAR points are generated from motion forecasting model, and with rich point features.

\noindent\textbf{Motion forecasting.} 
Given past observations of objects in a dynamic scene, motion forecasting aims to predict future trajectories of the objects.
Current state-of-the-art methods~\cite{ngiam2021scene,gu2021densetnt,liu2021multimodal,zeng2021lanercnn,ye2022dcms,varadarajan2022multipath++,nayakanti2022wayformer} learn the complex and nuanced interactions from data through deep neural networks.
Some other methods study joint 3D object detection and motion forecasting~\cite{luo2018fast, bansal2018chauffeurnet, casas2018intentnet, liang2020pnpnet, wu2020motionnet, zhang2020stinet, casas2021mp3, peri2022forecasting}, where detection can be an intermediate task. Among them, Fast and Furious~\cite{luo2018fast} relates to our work as they used motion forecasting to improve detection. In their tracklet decoding module, they aggregate motion forecasting and detector boxes through direct box averaging, which can be considered as a late fusion of motion forecasting and detection results.
Compared to them, this paper proposes an early fusion approach to leverage both motion forecasting (as MoDAR points) and LiDAR data and demonstrate superior performance to the late fusion alternative.

\section{Using MoDAR for 3D Object Detection}

In this section, we first introduce how to produce the virtual modality MoDAR, and then discuss how to fuse this new modality with a LiDAR point cloud for 3D object detection. \cref{fig:pipeline} illustrates the entire pipeline.

\subsection{MoDAR Point Generation} \label{sec:generate_modar}
We propose a virtual sensor modality, MoDAR, which represents object information propagated from past (or/and from future, in an offline setting) to the current frame. As shown in \cref{fig:pipeline}, MoDAR points at $T=0$ are generated from motion forecasting on a set of history subsequences (in the online setting) or/and a set of future subsequences (in the offline setting). Specifically, given a history subsequence of frames $T=-m-K:-m$ ($K+1$ context frames), forward motion forecasting predicts future trajectories of all detected objects at frame $T=-m$. We pick predictions that are $m$ frames into the future as the MoDAR points at the current frame. A MoDAR point's $XYZ$ coordinates are the predicted object center locations while the point features can include object bounding box size, heading, semantic class as well as other metadata from the motion forecasting model (e.g. confidence scores). In an offline setting, we can access future sensor data, which allows us to take a future subsequence and run reverse motion forecasting to predict object locations backwards.

\noindent\textbf{Motion forecasting (MF).}
To predict object locations into the future (or past), the motion forecasting (MF) involves three typical steps as shown in the bottom right of \cref{fig:pipeline}: detection, tracking, and prediction. We firstly run a pre-trained LiDAR based 3D object detector to localize and classify objects at every frame in the subsequence (results can be cached for overlapping subsequences), then we run multi-object tracking using a Kalman-filter based tracker~\cite{weng2019baseline}. Finally, a trajectory prediction model takes the tracked object boxes and predicts future object locations and headings. The trajectory prediction (or motion forecasting model) can be as simple as a constant velocity model~\cite{ettinger2021large}. It can also be a pre-trained deep network model such as MultiPath++~\cite{varadarajan2022multipath++} (MP++), which is more accurate especially for moving agents in a complex scene. Note that although the motion forecasting output is later used for detection, there is no cyclic dependency of the LiDAR-MoDAR detector and the motion forecasting, as we use a separate pre-trained LiDAR detector to generate the motion forecasting input. It is possible to share the same detector for LiDAR-MoDAR detection and motion forecasting input but requires iterative re-training to converge the detector and motion forecasting model -- see supplementary for more discussion and results.

\noindent\textbf{Extensions beyond a single prediction.}
To make our proposed MoDAR virtual modality more informative, we propose another two extensions. First, to fully leverage the point cloud sequence data, we can combine motion forecasting from separate history (or/and future) subsequences by taking a union of the MoDAR points generated from each subsequence. To distinguish their sources, we add an extra channel of the closet frame timestamp (in the subsequence) to the current frame.
Second, given an object track, data-driven motion forecasting models can predict several future trajectories, to handle the uncertainty in object behaviors. For example, MultiPath++~\cite{varadarajan2022multipath++} predicts 6 possible trajectories with different confidence scores. MoDAR can include all these predictions. To distinguish them, the trajectory confidence can be added as an addition field of the a MoDAR point.

\subsection{LiDAR-MoDAR 3D Object Detection}
The generated MoDAR points can be combined with LiDAR points at the current frame (or from a short time window around the current frame) for LiDAR-MoDAR multi-modal 3D object detection.
Since MoDAR is based on motion forecasting, it provides less accurate information than LiDARs for areas with good visibility. Therefore a MoDAR-only detection model would have unfavorable detection quality. However, we observe that MoDAR can provide complementary information to the LiDAR sensor especially when LiDAR points are sparse (long-range) or when objects are occluded. For example, when it is hard to estimate an object's size and heading when there are very few points, MoDAR points can help provide such information propagated from history (or future frames). When an object becomes occluded, the motion forecasting can still generate a virtual MoDAR point at the occluded region.

To leverage both LiDAR and MoDAR, we use an early fusion at the input level, for two reasons: First, compared to feature level fusion that often requires non-trival detector architecture update, early fusion is more flexible and can be easily adapted to nearly any off-the-shelf 3D object detectors; Second, compared to late fusion, early fusion is more effective in combining the complementary information from MoDARs and LiDARs. Besides early fusion, we find that adding another late fusion on top of it can further improve pedestrian detection, see \cref{sec:fusion} for more discussion and results.

As MoDAR points are light-weight, we can use many more context frames for our MoDAR-LiDAR detector than alternative methods that rely on point cloud based temporal data fusion. Compared to the number of LiDAR points in a single frame (around 200K in a frame from the Waymo Open Dataset~\cite{sun2020scalability}), the number of MoDAR points is marginal. There are $(N \times J)$ MoDAR points from one motion forecasting prediction, where $N$ is the number of objects (usually less than 100), and $J$ is the number of trajectories for each object (e.g. 6). Therefore, to representing information from one frame, MoDAR is around 300$\times$ more efficient than LiDAR. Due to its efficiency, MoDAR helps to include information from more context frames --- we use up to 180 frames (18 seconds: 9 seconds in history and 9 seconds in future) in our experiments.

\section{Experiments}
We evaluate LiDAR and MoDAR fusion detectors on the Waymo Open Dataset (WOD)~\cite{sun2020scalability}, a large scale autonomous driving dataset with challenging measurements covering different visibility levels. It contains 798 training sequences and 202 validation sequences. Each sequence is around 20 seconds (with around 200 frames at 10Hz).

We evaluate and compare methods with the recommended metrics, Average Precision (AP) and Average Precision weighted by Heading (APH), and report the results on both LEVEL\_1 (L1, easy only) and LEVEL\_2 (L2, easy and hard) difficulty levels for both vehicles and pedestrians.

\begin{table*}[tbh!]
\small
\begin{center}
\begin{tabular}{l|cc|cccc|cccc|l}
\toprule
\multirow{2}{*}{Model} & Frame & Offline & \multicolumn{2}{c}{Veh. L1 3D} & \multicolumn{2}{c|}{Veh. L2 3D} & \multicolumn{2}{c}{Ped. L1 3D} & \multicolumn{2}{c|}{Ped. L2 3D} & \multirow{2}{*}{\textbf{L2 3D mAPH}}  \\
  & [-p, +f] & Method? & AP  & APH & AP & APH & AP  & APH & AP & APH & \\\midrule
3D-MAN~\cite{yang20213d} & [-15, ~~0] &  & 74.5 & 74.0 & 67.6 & 67.1 & 71.7 & 67.7 & 62.6 & 59.0 & 63.1 \\
MPPNet~\cite{chen2022mppnet} & [~~-3, ~~0] & & 81.5 & 81.1 & 74.1 & 73.6 & 84.6 & 81.9 & 77.2 & 74.7 & 74.2 \\
MPPNet~\cite{chen2022mppnet} & [-15, ~~0] & & 82.7 & 82.3 & 75.4 & 75.0 & 84.7 & 82.3 & 77.4 & 75.1 & 75.1 \\
MVF++~\cite{qi2021offboard}$\ssymbol{2}$ & [~~-4, ~~0] &  & 79.7 & - & - &  -& 81.8 & -& -& -& -\\
3DAL~\cite{qi2021offboard} & [-$\infty$, $\infty$] & \cmark & \textbf{84.5} & \textbf{84.0} & 75.8 & 75.3 & 82.9 & 79.8 & 73.6 & 70.8 & 73.1 \\

\midrule \midrule
CenterPoint\cite{yin2021center}$\ssymbol{1}$ & [~~~0, ~~0] & & 72.9 & 72.3 & 64.7 & 64.2 & 71.9 & 58.3 & 64.3 & 51.9 & 58.1 \\
\footnotesize \: +\ourproposedmethod{} & [-91, ~~0] & & 76.1 & 75.6 & 68.9 & 68.4 & 73.8 & 68.7 & 66.9 & 62.1 & 65.3\improves{+7.2} \\
\footnotesize \: +\ourproposedmethod{} & [-91, 91] & \cmark & 80.1 & 79.5 & 73.7 & 73.2 & 76.4 & 71.4 & 69.9 & 65.0 & 69.2\improves{+11.1}\\
\midrule %
SWFormer\cite{sun2022swformer}$\ssymbol{1}$ & [~~~0, ~~0] & & 77.0 & 76.5 & 68.3 & 67.9 & 80.9 & 72.3 & 72.3 & 64.4 & 66.2 \\
\footnotesize \: +\ourproposedmethod{} & [-91, ~~0] &  & 80.6 & 80.1 & 72.8 & 72.3 & 83.5 & 79.5 & 75.7 & 71.8 & 72.1\improves{+5.9} \\
\footnotesize \: +\ourproposedmethod{} & [-91, 91] & \cmark & 82.9 & 82.3 & 75.6 & 75.1 & 85.2 & 81.3 & 78.0 & 74.3 & 74.7\improves{+8.5} \\
\midrule %
SWFormer\cite{sun2022swformer}$\ssymbol{1}$ & [~~-2, ~~0] & & 78.5 & 78.1 & 70.1 & 69.7 & 82.0 & 78.1 & 73.8 & 70.1 & 69.9 \\
\footnotesize \: +\ourproposedmethod{} & [-91, ~~0] & & 81.0 & 80.5 & 73.4 & 72.9 & 83.5 & 79.4 & 76.1 & 72.1 & 72.5\improves{+2.6}\\
\footnotesize \: +\ourproposedmethod{} & [-91, 91] & \cmark & \textbf{84.5} & \textbf{84.0} & \textbf{77.5} & \textbf{77.0} & \textbf{86.3} & \textbf{82.5} & \textbf{79.5} & \textbf{75.8}  & \textbf{76.4}\improves{+6.5}\\

\bottomrule %

\end{tabular}
\end{center}
\vskip -3ex
\caption{\textbf{3D object detection results on the WOD val set.} Complementary to LiDAR, our proposed virtual modality MoDAR significantly improves state-of-the-art 3D object detection models, CenterPoint and SWFormer. Our proposed method achieves state-of-the-art compared to previous methods. The Frame column illustrates the indices of the frames that are used for detection. We also mark a method as offline if it uses information from the future.
$\ssymbol{2}$: ensemble with 10 times test-time-augmentation. $\ssymbol{1}$: our re-implementation.
}
\label{tab:exp:main}
\vskip -4ex
\end{table*}

\subsection{Implementation Details} \label{sec:exp:detail}
\noindent\textbf{Generating MoDAR points.} To generate the proposed virtual modality MoDAR, we need to prepare (1) a detection and a tracking model to recognize objects from past (and future) point cloud sequences, and (2) a motion forecasting model to predict the future (and past) trajectories.

To prepare the training data for the motion forecasting model, we train LiDAR-only detectors (CenterPoint~\cite{yin2021center} or SWFormer~\cite{sun2022swformer}) on the Waymo Open Dataset train set and then run inference to get detection results on all frames at both train set and validation set frames. Then, to get object tracks, we use a simple Kalman Filter as the multi-object tracker to associate detection results across frames. Finally we apply a data-driven motion forecasting model MultiPath++~\cite{varadarajan2022multipath++} on the object tracks to predict their future (and past) trajectories.

The MultiPath++~\cite{varadarajan2022multipath++} is trained on the Waymo Open Motion Dataset (WOMD)~\cite{ettinger2021large} train set which has more than 70K sequences. Roadgraphs are not used because they are not available in WOD for inference. MultiPath++ takes tracked object boxes from 10 past frames and 1 current frames as input, and predicts object trajectories for the future 80 frames. To run inference on WOD, we pad and re-segment each WOD sequence to 91 frames in an overlapping manner. Therefore, given a 200 frame WOD sequence, we take each frame as a current frame to construct a 91-frame segment, obtaining 200 91-frame segments. Instead of using the original track sampling strategy in WOMD, we use all tracks during both training and testing.
We train two models: a forward MultiPath++ that takes the past tracks to predict the future, and a backward MultiPath++ that takes the future tracks to predict the past. Note that the backward model is only used for the offline setting, while the forward model is used for both online and offline models. On the WOD val set, the 8 second Average Displacement Error (ADE)~\cite{ettinger2021large} are 1.17 and 1.11 for the forward model and the backward model, respectively (see supplementary for more details).

\noindent\textbf{Fusing MoDAR and LiDAR.} We will firstly introduce the details of the MoDAR points, and then introduce how we fuse MoDAR and LiDAR together. 

A MoDAR point is structurally similar to a LiDAR point, including a 3D point coordinate and its feature. Specifically, MoDAR points are placed at the center of the predicted object location, and its feature has 13 channels that including object size (normalized by prec-omputed mean and std values), heading (represented by a unit vector), class (one hot encoding with depth 3), object tracking score (\ie, the average of object detection scores over 11 past/current frames), trajectory score, trajectory standard deviation (normalized by its mean and std values), and the timestamp of the closest frame in the input track. In the offline setting, the MoDAR points for a current frame are generated from 160 motion predictions (80 for future prediction, 80 for past prediction) that take different 11-frame input tracks. Therefore, we use the information from 181 frames. In the online setting, the MoDAR points for a current frame are from 80 motion predictions. Besides, the motion forecasting model, MultiPath++~\cite{varadarajan2022multipath++}, predicts 6 trajectories for each input track.

When fusing MoDAR with LiDAR, we first pad the LiDAR features and MoDAR features to the same length, and then add an additional field (0 for LiDAR and 1 for MoDAR) to indicate the modality of a point.

\begin{table}[t!]
\small
\begin{center}
\begin{tabular}{l|cc|ccc}
\toprule
\multirow{2}{*}{Predictor} & \multicolumn{2}{c|}{L2 APH} & \multicolumn{3}{c}{Veh. L2 APH} \\
  & Veh. & Ped. & STN & FST & VFST \\\midrule
- & 64.2 & 51.9 & 60.5 & 73.3 & 79.9 \\\midrule
Stationary & 73.0 & 62.1 & 72.0 & 73.2 & 77.6  \\
Constant Velocity & 69.2 & 61.4 & 67.5 & 74.6 & 79.0 \\
1 traj. MP++~\cite{varadarajan2022multipath++} & \textbf{73.5} & 64.4 & \textbf{71.4} & 75.5 & \textbf{82.3} \\
6 traj. MP++~\cite{varadarajan2022multipath++} & 73.2 & \textbf{65.0} & 70.5 & \textbf{76.2} & 81.3 \\

\bottomrule

\end{tabular}
\end{center}
\vskip -3ex
\caption{\textbf{Effects of motion forecasting model choices.} The metrics are vehicle L2 3D APH on the WOD val set. The first row is the CenterPoint detector using LiDAR data only. The other four rows are the same detector using LiDAR and MoDAR points (with different trajectory predictors). 
Results are also broken down by object speed (STN: stationary. FST: fast. VFST: very fast).
}
\label{tab:exp:predictor}
\vskip -4ex
\end{table}

\noindent\textbf{Detection Models.} We re-implemented two popular 3D point cloud detection models, the convolution-based CenterPoint~\cite{yin2021center} and the transformer-based SWFormer~\cite{sun2022swformer}. For CenterPoint, we train 160k steps with a total batch size of 64. For SWFormer, we train 80k steps with a total batch size of 256. The fusion of both LiDAR and MoDAR points are fed into these two models. During training, we apply data augmentations to both LiDAR and MoDAR points. % See supplementary for more details.

\subsection{Main Results}
\cref{tab:exp:main} shows adding MoDAR points can improve off-the-shelf 3D object detectors and compares our LiDAR-MoDAR detectors with prior art methods.

From the bottom half of \cref{tab:exp:main}, we see adding MoDAR points to two popular and powerful 3D detectors, CenterPoint~\cite{yin2021center} and SWFormer~\cite{sun2022swformer}, leads to significant gains across all metrics for both vehicle and pedestrian detection and in both online and offline settings.
For example, adding MoDAR points to the 1-frame CenterPoint base detector, we see $13.1$ L2 APH improvement (from $51.9$ to $65.0$) on pedestrians and $9.0$ L2 APH gains (from $64.2$ to $73.2$) on vehicles. The large gains apply to more powerful base detectors too. For the 3-frame SWFormer detector, adding MoDAR points can still lead to $7.3$ L2 APH improvement (from $69.7$ to $77.0$) for vehicles, and $5.7$ L2 APH improvement (from $70.1$ to $75.8$) for pedestrian.

\begin{figure*}[t]
\centering
\includegraphics[width=0.9\linewidth]{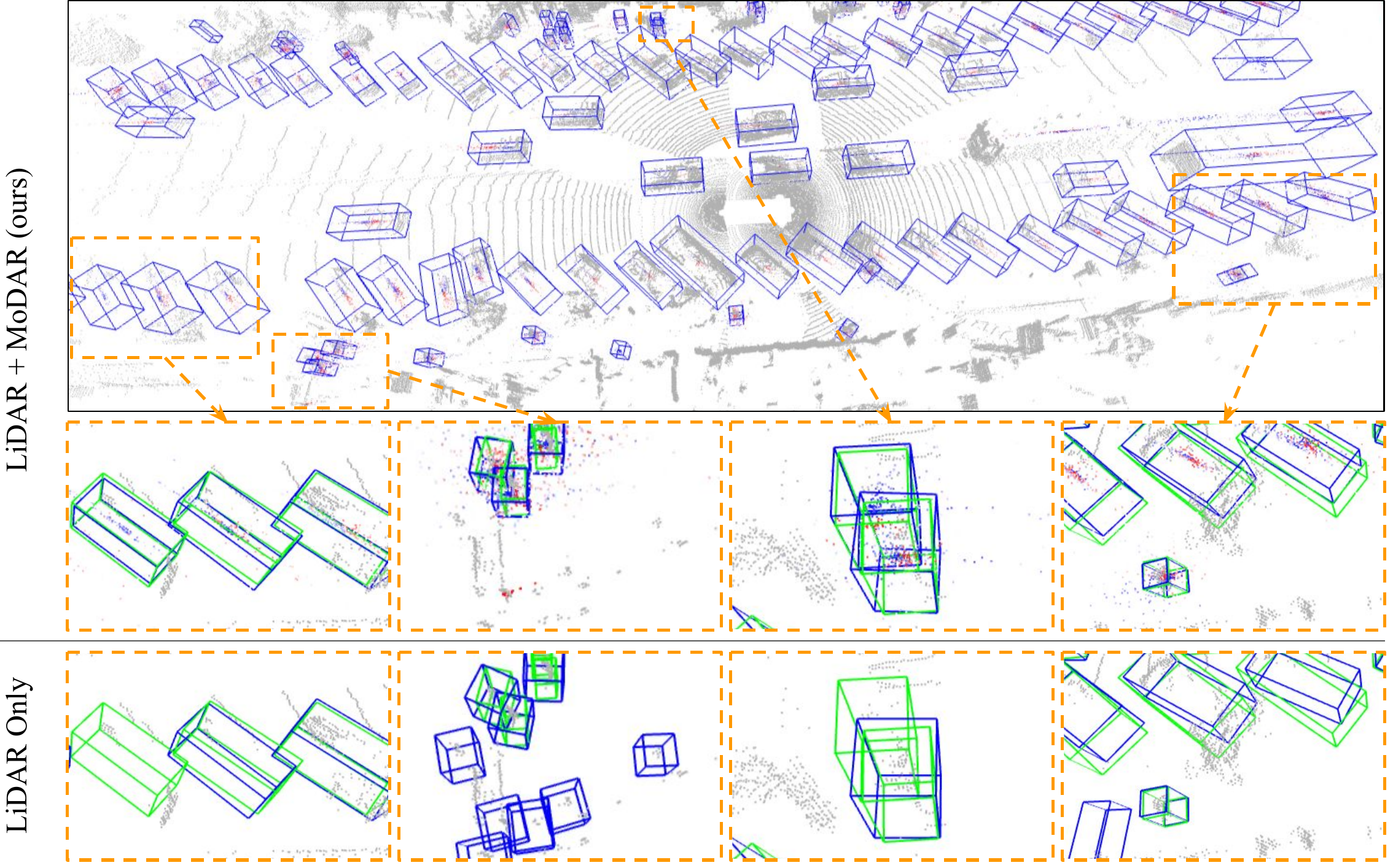}
\vskip -1ex
\caption{\textbf{Qualitative results of 3D object detection on the WOD val set.} \textcolor{blue}{Blue} boxes: model predictions; \textcolor{green}{Green} boxes: ground truth boxes. We used a 3-frame SWFormer as the base detector architecture and used MoDAR points generated from 180 context frames (offline setting). Comparing the results from LiDAR-MoDAR multi-modal detector versus the LiDAR only detector, we can see that the LiDAR-MoDAR detector can recognize more heavily occluded objects or estimate their shapes and headings more accurately.} 
\label{fig:exp:vis_rst}
\vskip -3ex
\end{figure*}

The improvement on L2 metric is more significant than the L1 metric. For example, MoDAR improves the 3-frame SWFormer by $7.3$ L2 APH and $5.9$ L1 APH (for vehicle), and by $5.7$ L2 APH and $4.4$ L1 APH (for pedestrian). Since L1 only considers relatively easy objects (usually more than 5 LiDAR points on them) while L2 considers all objects, this shows that MoDAR helps more in detecting difficult objects with low visibility (more breakdowns in \cref{sec:exp:range_breakdown} and visualizations in \cref{fig:exp:vis_rst}).

\cref{tab:exp:main} also compares our LiDAR-MoDAR detectors with prior art methods that leverage point cloud sequences for online/offline 3D object detection. Our method based on the 3-frame SWFormer gets the best mAPH results among all methods and achieves state-of-the-art numbers on all vehicle and pedestrian metrics. Note that although 3D Auto Labeling (3DAL) from Qi et al.~\cite{qi2021offboard} uses a stronger base detector (MVF++ with 5-frame input and test time augmentation) than our 3-frame SWFormer base detector, we can still achieve on par or stronger results than it with the extra input from MoDAR. In appendix, we demonstrate MoDAR can further improve a stronger baseline, LidarAug~\cite{leng2022lidaraugment}.

\subsection{Ablations and Analysis Experiments}
% This section discusses several designs to deep understand the proposed virtual modality MoDAR, and its behavior for the detection task.
This section ablates the MoDAR design and provides more analysis results.
Unless otherwise specified, all experiments in this section are based on the 1-frame CenterPoint detector using predictions from past and future 160 frames in total, and using early LiDAR-MoDAR fusion.

\subsubsection{Effects of motion forecasting models}
In \cref{tab:exp:predictor}, we compare detection results using MoDAR points generated from different motion forecasting models.
We ablate 4 different motion forecasting models. (1) Stationary predictor: it aggressively assumes all objects are stationary, predicting objects' future positions as their most recent positions. (2) Constant velocity predictor: it assumes all objects are moving at the constant velocity estimated from the observed frames. (3) MultiPath++~\cite{varadarajan2022multipath++} (MP++) predicting the most confident trajectory (1 traj.); (4) MultiPath++ predicting the top 6 confident trajectories (6 traj.). 

We see that MoDAR improves the CenterPoint baseline with all four predictors. The data-driven MultiPath++ model shows the best overall performance compared to other predictors.
% For example, the 6-trajectory MP++ achieves 65.0 L2 APH (ped), which is better than stationary predictor (62.1) and constant velocity (61.4).
When looking into the velocity breakdown metrics (provided by WOD), we observe that stationary predictor achieves the best performance ($72.0$ L2 APH) for the stationary (STN) vehicles, regresses on very fast (VFST) vehicles (from 79.9 to $77.6$). The constant velocity model is better than stationary predictor for fast (FST) and very fast (VFST) objects.
Note that the constant velocity model does not perform as well as the stationary predictor because the input track is noisy: even though the objects are not moving, detection noises can lead to wrong velocity estimation. Finally, the MP++ predictors perform the best for moving (\ie, fast and very fast) vehicles. 1-trajectory and 6-trajectory MP++ models lead to similar detection results. Note that we use 6-trajectory MP++ model as our final version. To seek higher model efficiency, 1-trajectory MP++ predictor can be selected which only includes $1/6$ MoDAR points compared to 6-trajectory MP++ model.

\subsubsection{Effects of different MoDAR point features}
\cref{tab:exp:info} ablates the importance of different object states in MoDAR points. Most information, includes object location, size, heading, type, confidence scores (from both tracking and motion forecasting), help improve the detection quality. By taking a closer look, we observe that the most important information is the object location, which improves $5.5$ L2 APH (from $64.2$ to $69.7$) for vehicles and $10.7$ L2 APH (from $51.9$ to $61.2$) for pedestrians. Since vehicle heading is relatively easy to estimate, adding headings to MoDAR features mainly helps pedestrian detection. For our main results, we used all features to get the best performance.

\begin{table}[t]
\vskip -0.15ex
\begin{center}
\resizebox{\columnwidth}{!}{

\begin{tabular}{ccccc|HcHc}
\toprule
\multirow{2}{*}{Location} & \multirow{2}{*}{Size} & \multirow{2}{*}{Heading} & \multirow{2}{*}{Class} & \multirow{2}{*}{Scores} & \multicolumn{2}{c}{Veh. L2} & \multicolumn{2}{c}{Ped. L2} \\
  &    &  &  &  & AP  & APH & AP & APH  \\\midrule
\xmark & \xmark & \xmark & \xmark & \xmark & 64.7 & 64.2 & 64.3 & 51.9 \\ \midrule
\cmark & \xmark & \xmark & \xmark & \xmark & 70.3 & 69.7 & 67.0 & 61.2 \\ \midrule
\cmark & \cmark & \xmark & \xmark & \xmark & 71.9 & 71.2 & 68.3 & 62.7 \\ \midrule
\cmark & \cmark & \cmark & \xmark & \xmark & 71.5 & 70.9 & 67.8 & 64.0 \\ \midrule
\cmark & \cmark & \cmark & \cmark & \xmark & 71.9 & 71.3 & 68.1 & 64.1 \\ \midrule
\cmark & \cmark & \cmark & \cmark & \cmark & \textbf{73.7} & \textbf{73.2} & \textbf{69.9} & \textbf{65.0} \\ \bottomrule

\end{tabular}}
\end{center}
\vskip -4ex
\caption{\textbf{Effects of different object state features in MoDAR points.} The metric is L2 3D APH on the WOD val set.}
\label{tab:exp:info}
\vskip -4ex
\end{table}

\begin{figure}[b]
\vskip -4ex
\centering
\includegraphics[width=\linewidth]{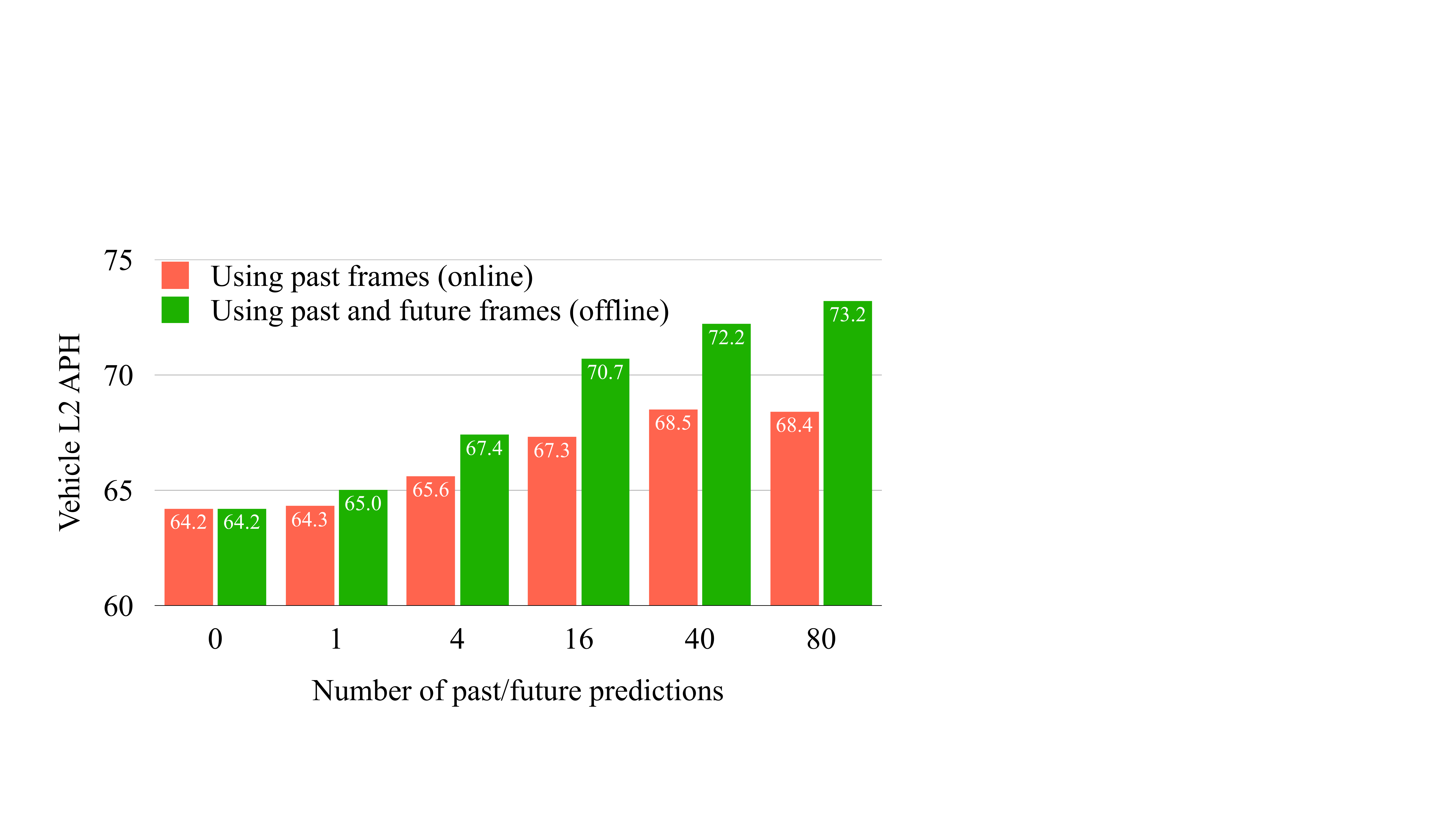}
\vskip -1ex
\caption{\textbf{Effects of MoDAR temporal context sizes on 3D object detection.} The metrics are vehicle L2 3D APH on the WOD val set. The (left) red bars are the performance of models using only MoDAR points generated from the past frames, while the (right) green bars are the performance of models that use the same number of past and future frames for MoDAR point generation.
} 
\label{fig:exp:num_frames}
\vskip -0.75ex
\end{figure}

\subsubsection{Are long-term point cloud sequences helpful?}
In \cref{fig:exp:num_frames}, we show the impact of the temporal ranges (what frames are used for motion forecasting) of MoDAR points on detection.
We split the study to two settings: online and offline. In the online setting, only past frames are used to generate MoDAR points. For the offline setting, both past and future frames are used. We include the same number of future frames as past frames for the offline setting. For the cases of using K past predictions, we select past subsequences from T=-11:-1 to T=-K=10:-K (K subsequences) and use the forward motion forecasting from them to generate MoDAR points. Similarly for the reverse prediction from future subsequences.

The results are shown in \cref{fig:exp:num_frames}. The red bars are the performance of the online setting, while the green bars are for the offline setting. We observe that for both setting, adding MoDARs from more past or future predictions generally lead to better detection and this improvement does not saturate until using MoDARs from 80 past and 80 future predictions. It is also noteworthy that the future frames provide unique information that significantly improves the results compared to only using past frames.

\begin{table}[t!]
\small
\begin{center}
\begin{tabular}{l|ccc|c}
\toprule
Model & 0-30m & 30-50m & 50m+ & All \\\midrule
LiDAR only & 90.4 & 69.7 & 45.6 & 69.7 \\
LiDAR + MoDAR & \textbf{92.2} & \textbf{76.9} & \textbf{58.5} & \textbf{77.0} \\
& \improves{+1.2} & \improves{+7.2} & \improves{+12.9} & \improves{+7.3} \\

\bottomrule

\end{tabular}
\end{center}
\vskip -4ex
\caption{\textbf{Distance breakdown for LiDAR-based and LiDAR-MoDAR based detection.} The metrics are vehicle L2 3D APH on the WOD val set across different ground-truth depth ranges. The base detector is a 3-frame SWFormer.
}
\vskip -4ex
\label{tab:exp:range_breakdown}
\end{table}

\subsubsection{Performance breakdown by object distances} \label{sec:exp:range_breakdown}
To better understand how MoDAR improves the LiDAR-based 3D object detection models, we provide both qualitative and quantitative analysis based on the 3-frame SWFormer model and our MoDAR variant. 

Following previous works~\cite{qi2021offboard,li2022deepfusion}, we divide the vehicles into three groups based on their distance to the ego-car: within 30 meters (short-range), from 30 to 50 meters (mid-range), and beyond 50 meters (long-range). \cref{tab:exp:range_breakdown} shows the \textit{relative} gains by using MoDAR. MoDAR improves the results in all distance ranges. In particular, it achieves a much more significant gains for long-range vehicles (by $12.9$ APH, $28.3\%$ relatively) than short-range vehicles (by $1.8$ APH, $2.0\%$ relatively). This is likely because long-range objects have very sparse points in their observations, making it difficult to estimate their locations, headings and sizes. MoDAR fills this gap to a large extent.

\subsubsection{Comparing LiDAR-MoDAR fusion methods}
\label{sec:fusion}
In \cref{tab:exp:fusion}, we compare detection results using a single modality and results using both LiDAR and MoDAR modalities using different fusion strategies.

For LiDAR-only detection, we train a 3-frame SWFormer that only takes LiDAR points as input. For MoDAR-only, we directly use the motion forecasted boxes as the detection output (assuming constant box sizes and box elevation). Note that motion predictions from nearby frames (e.g. 1 or 2 frames away) can give very similar results as to detection from the current frame, as the scene does not change dramatically between nearby frames. With some hyper parameter tuning, we select MoDAR points from the closest 10 predictions (5 past and 5 future) for the MoDAR-only detection and then apply a weighted 3D box fusion~\cite{solovyev2021weighted} to aggregate overlapping boxes (see supplementary for more details and ablations). From \cref{tab:exp:fusion} first two rows, we can see that LiDAR-based detection gets more accurate results than MoDAR-based ones especially for pedestrians, for which motion forecasting can be noisy.

Fast and Furious~\cite{luo2018fast} used a late fusion approach to combine detector and motion forecasting results through box averaging. To implement a late fusion method, we compute weighted box averaging~\cite{solovyev2021weighted} of boxes from current frame LiDAR detection and motion predictions from nearby 10 predictions (past 5 and future 5). For the early fusion, we use motion forecasting from 160 predictions (past 80 and future 80) to generate MoDAR points. In \cref{tab:exp:fusion} third and fourth rows, we can see that early fusion achieves significantly better results than the late fusion.
In the last row, we show that if we combine the early and late fusion by fusing forecasted boxes from nearby 10 frames with LiDAR-MoDAR detection, we can further improve detection quality. For our main results in \cref{tab:exp:main}, we take advantages of late fusion for pedestrian detection (early+late fusion).

\section{Conclusions}
\begin{table}[t]
\vskip 0.75ex
\small
\begin{center}
\resizebox{\columnwidth}{!}{
\begin{tabular}{ccc|cc|cc}
\toprule
\multirow{2}{*}{LiDAR} & \multirow{2}{*}{MoDAR} & Fusion & \multicolumn{2}{c|}{Veh. L2} & \multicolumn{2}{c}{Ped. L2} \\
 & & Method & AP & APH & AP & APH \\\midrule
\cmark & \xmark & - & 70.1 & 69.7 & 73.8 & 70.1 \\
\xmark & \cmark & - & 67.4 & 66.8 & 69.6 & 63.8 \\\midrule
\cmark & \cmark & Early &77.5 & 77.0 & 77.8 & 74.4 \\
\cmark & \cmark & Late & 70.9 & 70.4 & 76.4 & 72.3 \\
\cmark & \cmark & Early+Late & \textbf{77.6} & \textbf{77.1} & \textbf{79.5} & \textbf{75.8} \\
  
\bottomrule

\end{tabular}
}
\end{center}
\vskip -4ex
\caption{\textbf{Compare detection results with different modalities and different fusion methods.} The metrics are 3D and BEV L2 APH on the WOD val set. We use a 3-frame SWFormer.
% \todo{Fill in the numbers.}
}
\vskip -4ex
\label{tab:exp:fusion}
\end{table}

In this paper, we proposed MoDAR, a virtual sensor modality that uses motion forecasting to propagate object states from past and future frames to a target frame. Each MoDAR point represents a prediction of an object's location and states on a forecasted trajectory. The MoDAR points generated from a point cloud sequence can be fused with other sensor modalities such as LiDAR to achieve more robust 3D object detection especially for cases with low visibility (occluded) or in long range. Due to its simplicity, the MoDAR idea can be applied to a wide range of existing detectors not even restricted to point-cloud-based ones. Evaluated on the Waymo Open Dataset, we have demonstrated the effectiveness of the MoDAR points for two popular 3D object detectors, achieving state-of-the-art results. We have also provided extensive analysis to understand different components of the MoDAR modality.

We believe this work provides another perspective of the relationship between detection and motion forecasting. In the future, it would be appealing to study how to jointly optimize motion prediction and detection, as well as revisiting the interface design between them.

{\small{
\vskip 0.75ex
\noindent\textbf{Acknowledgement.}
We specially thank Yurong You for idea discussion and preliminary exploration, and thank Scott Ettinger, and Chiyu “Max” Jiang for insightful discussion and technical support. Yingwei Li thank Longlong Jing, Zhaoqi Leng, Pei Sun, Tong He, Mingxing Tan, Hubert Lin, Xuanyu Zhou, Mahyar Najibi, and Kan Chen for tutorial, discussion and technical support.}
}

%%%%%%%%% REFERENCES
{\small
\bibliographystyle{ieee_fullname}
\bibliography{egbib, yin_ref}
}

\clearpage

\appendix
\section*{Appendix}
\section{Motion Forecasting Model}
In \cref{sec:exp:detail}, we mentioned that a forward and a backward MultiPath++ are trained for generating MoDAR points. In this section, we provide more details about training and evaluating the MultiPath++ models.

\paragraph{Close the domain gap between WOMD and WOD.} When constructing Waymo Open Motion Dataset (WOMD) and training the motion forecasting models, people intentionally mines the interesting trajectories, such as the following pairwise cases: merges, lane changes, unprotected turns, intersection left turns, intersection right turns, pedestrian-vehicle intersections, cyclist-vehicle in intersections, intersections with close proximity, and intersections with high accelerations~\cite{ettinger2021large}. Different from WOMD, most trajectories in Waymo Open Dataset (WOD) are less interesting: cars are usually parked or moving with a constant velocity~\cite{sun2020scalability}.

Therefore, to close the trajectories sampling gap, when training MultiPath++~\cite{varadarajan2022multipath++} on WOMD, we change the original sampling strategy to a dense sampling strategy, which uses all tracks for training instead of sampling the interesting tracks. \cref{tab:supp:sampling} shows the performance comparison when training with different sampling strategies and testing on different dataset. When training with the original WOMD, the results are better on the original WOMD validation set. This is because both original WOMD training and validation sets sample the interesting trajectories. However, when training with the dense sampled WOMD, the results on WOD validation set is better. For example, the Average Displacement Error (ADE) is reduced from 1.83 to 1.17 on WOD validation set, by changing the original sampling strategy to the dense sampling strategy.

\begin{table*}[tb]
    \centering
    \begin{tabular}{ll|cccc}
    \toprule
    Training Set & Validation Set & ADE & FDE & minADE & minFDE  \\ 
    \midrule
    \multirow{2}{*}{\begin{tabular}[l]{@{}l@{}}WOMD \\ (Original)\end{tabular} } & WOMD Val. &  3.34 & 10.2 & 1.40 & 4.01 \\
     & WOD Val. & 1.83 & 9.22 & 0.82 & 3.67 \\ \midrule
    \multirow{2}{*}{\begin{tabular}[l]{@{}l@{}}WOMD \\ (Dense)\end{tabular}} & WOMD Val. & 3.61 & 11.1 & 1.42 & 3.74  \\
     & WOD Val. & 1.17 & 5.45 & 0.55 & 2.27 \\
    \bottomrule
    \end{tabular}
    \caption{Compare different sampling strategies when training the motion forecasting model, MultiPath++, on Waymo Open Motion Dataset (WOMD). We test the trained model on the validation set of both WOMD and WOD. We observe that the dense sampling strategy leads to lower error on WOD validation set. ADE, FDE, minADE, and minFDE are evaluation metrics (lower is better) for the motion forecasting task. }
    \label{tab:supp:sampling}
\end{table*}

\paragraph{Forward and Reverse Motion Forecasting Models.} Besides the past point cloud sequence, the offboard detection set up also takes the information from the future point cloud sequence. To propagate future object information to the current frame, we train a reverse motion forecasting model. Specifically, we prepared the reversed training set based on the WOMD, and also prepared the reversed WOD for generating MoDAR points. When preparing the training set, we resplit all 91 frame trajectories to 11 frame input track and 80 frame ground truth track as training label. Different from the forward dataset, the backward dataset take the last 11 frames as the input, and guide the model to predict the first 80 frame trajectories. Besides, we reverse the velocity vector of each object accordingly. When preparing the WOD inference set, instead of using the original timestamp $T_\textrm{original}$, we assign a virtual (negative) timestamp $T_\textrm{virtual}$ for each detection. The timestamp will be normalized before feeding into the motion forecasting models. After we re-assign the virtual timestamp to each detection box, we proceed the tracking and motion forecasting as forward counterpart. Finally, we convert the virtual timestamp back to the original timestamp by $T_\textrm{original} = - T_\textrm{virtual} + c$. Finally, we compare the forward and reverse motion forecasting model in \cref{tab:supp:reversebp}, showing that the reverse model is as good as (or even slightly better than) the forward model.

\begin{table*}[tb]
    \centering
    \begin{tabular}{l|cccc}
    \toprule
     & ADE & FDE & minADE & minFDE  \\ 
    \midrule
    Forward MP++ & 1.17 & 5.45 & 0.55 & 2.27 \\
    Reverse MP++ & 1.11 & 4.70 & 0.51 & 1.76 \\
    \bottomrule
    \end{tabular}
    \caption{Compare the performance of the forward and the reverse motion forecasting models. We observe that the reverse motion forecasting model is as good as (or even slightly better than) the forward one.}
    \label{tab:supp:reversebp}
\end{table*}

% \section{Possibility to share the same 3D detector for LiDAR-MoDAR detection and motion forecasting input.}
\section{Sharing 3D Detectors for LiDAR-MoDAR detection and Motion Forecasting}
As we illustrated in \cref{fig:pipeline}, our pipeline needs two 3D detection models: (1) a LiDAR 3D object detection model to prepare the input tracks for motion forecasting model, and (2) a LiDAR-MoDAR 3D object detection model for the final detection results. Although the architecture of these two models are the same, their weights are trained separately. In this section, we discuss the possibility to consolidate these two models,~\ie,~use a shared model with the same weights for both LiDAR-MoDAR detection and motion forecasting input. We explore the impact to performance if the LiDAR-MoDAR detector model takes the MoDAR points generated by itself (MoDAR from its detection boxes).

The results are shown in \cref{tab:supp:share}. We use the 1-frame SWFormer as the baseline model (\#W1), and use an in-house motion forecasting model that is slightly stronger than MultiPath++ reported in the main paper. We observe that when the MoDAR points are generated differently during training and validation, the performance will drop. For example, when training and evaluating with MoDAR points generated by \#W1, the L2 3D mAPH is 74.5. However, if evaluating this model with the MoDAR points generated by \#W2, the performance drops by 3.7 (from 74.5 to 70.8) L2 3D mAPH, even though the MoDAR points from \#W2 is more accurate than \#W1. We also observe that retraining the detector model again helps reduce this gap. Specifically, for model \#W3, when training with MoDAR points from \#W2 and evaluate with MoDAR points from \#W3, the performance only drops by 1.4 (from 74.8 to 73.4) L2 3D mAPH, which is smaller than the 3.7 L2 3D mAPH gap for model \#W2. Therefore, we hypothesize iterative training can potentially mitigate this problem. However iterative re-training would make the training process more complex. As a future work, we can explore other techniques (such as adding noise to MoDAR points during training, or generating MoDAR points on-the-fly) to improve the robustness of taking MoDAR points from different models.
% \ywli{Write more details of our experiments.}

\begin{table*}[tbh!]
\small
\begin{center}
\begin{tabular}{ll|cc|cccc|cccc|l}
\toprule
Model & \multirow{2}{*}{Model} & \multicolumn{2}{c|}{MP++ inputs} & \multicolumn{2}{c}{Veh. L1 3D} & \multicolumn{2}{c|}{Veh. L2 3D} & \multicolumn{2}{c}{Ped. L1 3D} & \multicolumn{2}{c|}{Ped. L2 3D} & \multirow{2}{*}{\textbf{L2 3D mAPH}}  \\
ID & & @train & @eval & AP  & APH & AP & APH & AP  & APH & AP & APH & \\\midrule
\#W1 & SWFormer\cite{sun2022swformer} & - & - & 77.0 & 76.5 & 68.3 & 67.9 & 80.9 & 72.3 & 72.3 & 64.4 & 66.2 \\
\midrule
\multirow{2}{*}{\#W2} & \multirow{2}{*}{\footnotesize \: +\ourproposedmethod{}} & \multirow{2}{*}{\#W1} & \#W1 & 83.2 & 82.6 & 75.9 & 75.3 & 84.0 & 80.5 & 76.7 & 73.7 & 74.5\improves{+8.3} \\
 & & & \#W2 & 80.2 & 79.6 & 73.2 & 72.6 & 80.0 & 76.0 & 72.7 & 69.0 & 70.8\improves{+4.6} \\
\midrule
\multirow{2}{*}{\#W3} & \multirow{2}{*}{\footnotesize \: +\ourproposedmethod{}} & \multirow{2}{*}{\#W2} & \#W2 & 83.6 & 83.0 & 76.4 & 75.9 & 84.4 & 80.8 & 77.1 & 73.7 & 74.8\improves{+8.6} \\
 & & & \#W3 & 82.3 & 81.7 & 75.3 & 74.8 & 82.6 & 79.0 & 75.3 & 71.9 & 73.4\improves{+7.2} \\

\bottomrule %

\end{tabular}
\end{center}
\caption{The performance comparison when generating MoDAR points by different models during evaluation. We observe that using different model to (feed to MP++ as inputs to) generate MoDAR points during training harms the final detection performance. Iterative training can mitigate this performance drop.}
\label{tab:supp:share}
\end{table*}

\begin{table*}[tb]
\small
\begin{center}
\begin{tabular}{l|cc|cccc|cccc|l}
\toprule
\multirow{2}{*}{Model} & Frame & Offline & \multicolumn{2}{c}{Veh. L1 3D} & \multicolumn{2}{c|}{Veh. L2 3D} & \multicolumn{2}{c}{Ped. L1 3D} & \multicolumn{2}{c|}{Ped. L2 3D} & \multirow{2}{*}{\textbf{L2 3D mAPH}}  \\
  & [-p, +f] & Method? & AP  & APH & AP & APH & AP  & APH & AP & APH & \\\midrule
MVF++~\cite{qi2021offboard}$\ssymbol{2}$ & [~~-4, ~~0] &  & 79.7 & - & - &  -& 81.8 & -& -& -& -\\
\footnotesize \: +3DAL~\cite{qi2021offboard} & [-$\infty$, $\infty$] & \cmark & 84.5 & 84.0 & 75.8 & 75.3 & 82.9 & 79.8 & 73.6 & 70.8 & 73.1 \\ \midrule
LidarAug~\cite{leng2022lidaraugment}$\ssymbol{1}$ & [~~-2, ~~0] & & 81.4 & 80.9 & 73.3 & 72.8 & 84.1 & 80.4 & 76.5 & 72.9 & 72.9 \\
\footnotesize \: +\ourproposedmethod{} & [-91, ~91] & \cmark & \textbf{86.3} & \textbf{85.8} & \textbf{79.5} & \textbf{79.0} & \textbf{87.7} & \textbf{84.6} & \textbf{81.1} & \textbf{78.0} & \textbf{78.5} \\
\bottomrule

\end{tabular}
\end{center}
\caption{MoDAR based on a stronger detection LidarAug~\cite{leng2022lidaraugment}.
$\ssymbol{2}$: ensemble with 10 times test-time-augmentation. $\ssymbol{1}$: our re-implementation.
}
\label{tab:StrongerBaseline}
\end{table*}

\section{Implementation Details of Detectors}
For CenterPoints and SWFormer LiDAR-only models, we apply data augmentations during training following the original SWFormer implementation~\cite{sun2022swformer}: randomly rotating the world by yaws, randomly flipping the world along y-axis, randomly scaling the world, and randomly dropping points. For the MoDAR-LiDAR fusion model, we first combine MoDAR and LiDAR points together, and then apply data augmentation to the fused point cloud. Note that these data augmentation only change the 3D coordinate of points, but keep the point feature unchanged.

\section{MoDAR-LiDAR Fusion}
\paragraph{Late fusion implementation details.} We implemented the MoDAR-LiDAR late fusion by a weighted box fusion strategy~\cite{solovyev2021weighted}. Since LiDAR signal shows better performance, we set the weight of the LiDAR predictions as 0.9 and set the weight of MoDAR predictions as 0.1. We finally keep top 300 boxes sorted by the confidence scores.

\paragraph{Fusing MoDAR from different frames.} In \cref{tab:exp:fusion}, we directly get the detection results from MoDAR. In this section, we introduce more details about how to generate detection boxes from MoDAR. As we mentioned, each MoDAR point represents a predicted 3D box. The location of the MoDAR point is the predicted center of the object, while the object size is stored in the MoDAR point feature. Therefore, we have a large number of 3D boxes predicted by different motion forecasting models. We also use the weighted box fusion strategy~\cite{solovyev2021weighted} to fuse these boxes together. Specifically, the boxes generated by recent predictors will have higher weights. Take the $5 \times 2$ predictions in \cref{tab:supp:fusion} as an example: we take the boxes from the closest 5 past and 5 future predictors, with the weight of $1.0$, $0.8$, $0.6$, $0.4$, and $0.2$. The results are shown in \cref{tab:supp:fusion}, and we call this method as late fusion because it is a box-level fusion strategy. We observe that using the closest 5 past and 5 future predictors achieves the best results. Fusing boxes from more predictors does not help because the long-term predictors predict less accurate boxes.

On the other hand, in this section, we also explore the early fusion strategy to fuse the MoDAR points from different predictors. Specifically, we put all MoDAR points (but no LiDAR points) as the input of the 3D object detection model. According to the results shown in \cref{tab:supp:fusion}, early fusion is more effective than the late fusion, and it can take the MoDAR points from more predictors even if the predictors are not close to the current frame. For example, our best MoDAR-only early fusion model achieves 70.5 Vehicle L2 APH and 72.0 Pedestrian L2 APH, which is already better than the LiDAR-only model with 69.7 Vehicle L2 APH and 70.1 Pedestrian L2 APH (shown in \cref{tab:exp:fusion} in the main paper).

\begin{table}[t!]
\small
\begin{center}
\resizebox{\columnwidth}{!}{
\begin{tabular}{Hcc|cc|cc}
\toprule
\multirow{2}{*}{LiDAR} & Number of & Fusion & \multicolumn{2}{c|}{Veh. L2} & \multicolumn{2}{c}{Ped. L2} \\
 & Predictions & Method & AP & APH & AP & APH \\\midrule
\xmark & 1 $\times$ 2 & Late & 65.6 & 65.0 & \textbf{69.6} & 63.7 \\
\xmark & 5 $\times$ 2 & Late & \textbf{67.4} & \textbf{66.8} & \textbf{69.6} & \textbf{63.8} \\
\xmark & 10 $\times$ 2 & Late & 67.1 & 66.5 & 62.9 & 57.6 \\
\xmark & 15 $\times$ 2 & Late & 66.1 & 65.6 & 52.5 & 48.2 \\ \midrule
\xmark & 5 $\times$ 2 & Early & 70.3 & 68.6 & 74.5 & 70.2 \\
\xmark & 10 $\times$ 2 & Early & 70.4 & 69.3 & 75.5 & 71.4 \\
\xmark & 20 $\times$ 2 & Early & \textbf{71.2} & \textbf{70.5} & \textbf{75.8} & \textbf{72.0} \\
\xmark & 40 $\times$ 2 & Early & 70.9 & 70.2 & 74.8 & 70.8 \\
\xmark & 80 $\times$ 2 & Early & 69.0 & 68.4 & 74.3 & 70.3 \\
\bottomrule

\end{tabular}
}
\end{center}
\caption{Fusing MoDAR from different predictors. We compare the early and the late fusion strategies, and explore to fuse different number of predictions ("$\times2$" means fusing the predictions from both past and future predictors).}
\label{tab:supp:fusion}
\end{table}

\section{Latency}
In this section, we compared the latency of our MoDAR-LiDAR fusion detection model with the LiDAR-only detection model, based on our re-implementation of the 3-frame SWFormer. We measure the latency using an in-house GPU. The average latency of our baseline 3-frame SWFormer is 172ms per frame. Note that this latency is considerably higher than the 20ms latency reported in the original SWFormer paper~\cite{sun2020scalability}, which is mainly because our research-oriented implementation is not optimized with respect to the fused transformer kernels~\cite{sun2020scalability} and the hardware devices are different. However, the comparisons below are under the same hardware devices and under the same implementation.

We measure the latency of three LiDAR-only models, the LiDAR-only SWFormer with 3-, 5-, or 7-frame LiDAR point cloud input, and our MoDAR-LiDAR fusion model that takes 3-frame LiDAR point cloud and MoDAR points from 160 predictors. The latency are shown in \cref{tab:supp:latency}. As we can see, the latency of our LiDAR-MoDAR fusion detector is between 3-frame and 5-frame LiDAR-only model, indicating the marginal computational complexity for using MoDAR points. Note that for the onboard system, we can cache the motion forecasting signal with little overhead, because motion forecasting is usually an important module of an autonomous driving system. 
For the offboard application, the latency of our motion forecasting model MultiPath++ is 217 ms, which is similar to the detection model. Compared with 3DAL~\cite{qi2021offboard} that takes 15min to process a 200-frame sequence, our offboard system takes about $221+172+217*2=827$ms to process a frame,~\ie, 3 minutes per 200-frame sequence, which is about 5$\times$ faster than 3DAL. As future work, by implementing customized kernels and optimizing network architectures, we expect to further reduce the latency.

\begin{table}[t!]
\small
\begin{center}
\begin{tabular}{cc|c}
\toprule
LiDAR & MoDAR & Latency (ms) \\
3 frames & \xmark & 172 \\
5 frames & \xmark & 247 \\
7 frames & \xmark & 276 \\ \midrule
3 frames & \cmark & 221 \\

\bottomrule

\end{tabular}

\end{center}
\caption{Latency comparison between LiDAR-MoDAR fusion and LiDAR-only models. The latency of LiDAR-MoDAR fusion model is between 3-frame and 5-frame LiDAR-only models.}
\label{tab:supp:latency}
\end{table}

\section{Results on the WOD Test Set}
\cref{tab:testset} shows vehicle and pedestrain detection results comparison with our baseline, SWFormer~\cite{sun2022swformer}. We observe a similar improvement compared with the results on validation set. This further indicates the effectiveness of our proposed method.
\begin{table}[tb]
\small
\setlength{\tabcolsep}{2.5pt}
\begin{tabular}{l|c|cc|cc}
\toprule
\multirow{2}{*}{Model} & mAPH & \multicolumn{2}{c|}{Veh AP/APH 3D} & \multicolumn{2}{c}{Ped AP/APH 3D}  \\
  & L2 & L1  & L2 & L1  & L2 \\\midrule
    3DAL & - & 85.8/85.5 & 77.2/76.9 & - & - \\
    SWFormer & 73.4 & 82.9/82.5 & 75.0/74.7 & 82.1/78.1 & 75.9/72.1 \\\midrule
    \footnotesize \: +\ourproposedmethod{} & \textbf{78.9} & \textbf{88.0}/\textbf{87.5} & \textbf{81.2}/\textbf{80.8} & \textbf{85.8}/\textbf{82.5} & \textbf{80.2}/\textbf{77.0} \\
\bottomrule
\end{tabular}
\caption{Compare WOD test set results with our baseline method, SWFormer~\cite{sun2022swformer}. mAPH/L2 is the offical ranking metric on the WOD leaderboard.}
\label{tab:testset}
\end{table}

\section{Generalizing to Stronger Detectors}
To show our method generalizes, we use LidarAug-SWFormer~\cite{leng2022lidaraugment} as a stronger baseline. Shown in \cref{tab:StrongerBaseline}, adding MoDAR leads to consistent gains and significantly outperforms previous methods. For example, we achieves 78.5 L2 3D mAPH, which is significantly better 3DAL by 5.4 L2 3D mAPH. 

\end{document}